\documentclass[10pt,twocolumn,letterpaper]{article}

\usepackage{url}
\usepackage{cvpr}
\usepackage{times}
\usepackage{epsfig}
\usepackage{graphicx}
\usepackage{amsmath}
\usepackage{amssymb}
\usepackage{multirow}
\usepackage{algorithm}  
\usepackage{algorithmicx}  
\usepackage{algpseudocode}  
\usepackage{amsmath}  
\usepackage{capt-of}

\usepackage{pifont}
\newcommand{\cmark}{\ding{51}}%
\newcommand{\xmark}{\ding{55}}%

\newcommand\blfootnote[1]{%
	\begingroup
	\renewcommand\thefootnote{}\footnote{#1}%
	\addtocounter{footnote}{-1}%
	\endgroup
}		

\usepackage[breaklinks=true,bookmarks=false,colorlinks]{hyperref}

\cvprfinalcopy 

\def\cvprPaperID{****} 

\ifcvprfinal\fi
\begin{document}

\title{3D Sketch-aware Semantic Scene Completion \\ via Semi-supervised Structure Prior}

\author{
	Xiaokang Chen$^{1*}$
	~~\quad Kwan-Yee Lin$^{2}$
	~~\quad Chen Qian$^{2}$
	~~\quad Gang Zeng$^{1\dagger}$  
	~~\quad Hongsheng Li$^3$	\vspace{0.1cm}\\
$^1$Key Laboratory of Machine Perception (MOE), School of EECS, Peking University ~~\quad\\
$^2$SenseTime Research ~~\quad  
$^3$The Chinese University of Hong Kong\\
}

\twocolumn[{
\renewcommand\twocolumn[1][]{#1}
\maketitle
\thispagestyle{empty} 
\begin{center}
    \centering
    \includegraphics[width=1.0\textwidth]{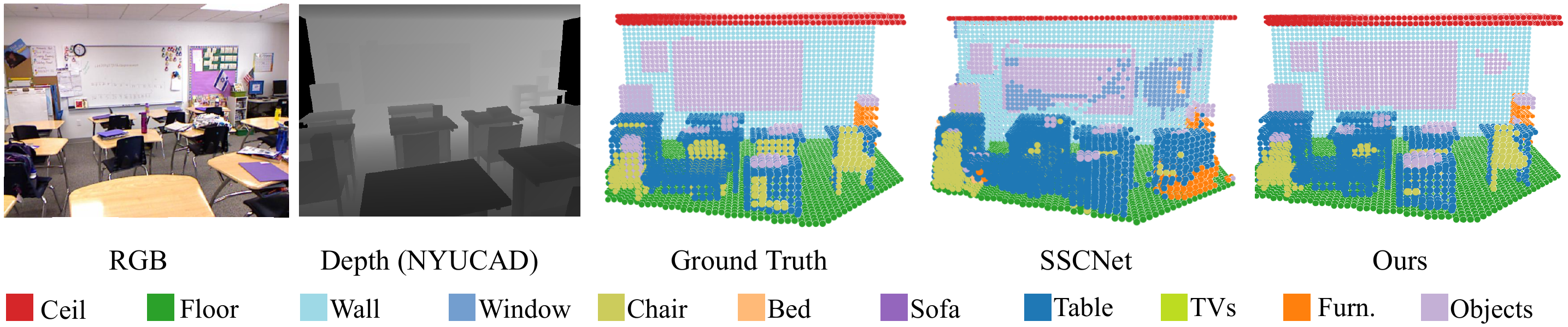}
    \captionof{figure}{\small{\textbf{Visualization of Semantic Scene Completion task.} From left to right: (1) RGB input, (2) depth map, (3) ground truth of semantic scene completion, (4) result of SSCNet~\cite{song2017semantic-sscnet}, (5) result of the proposed method. Our method generates a more reasonable result and obtains a better intra-class consistency and inter-class distinction compared with SSCNet~\cite{song2017semantic-sscnet}, a classic method that models context on implicitly embedded depth feature that learnt from general 3D CNNs.}}
    \label{fig:intro}
\end{center}
}]

\blfootnote{
	* This work was done during an internship at SenseTime Research. 
	
\hspace{0.24cm} $\dagger$ Gang Zeng is the corresponding author.
}

\begin{abstract}
The goal of the Semantic Scene Completion (SSC) task is to simultaneously predict a completed 3D voxel representation of volumetric occupancy and semantic labels of objects in the scene from a single-view observation. Since the computational cost generally increases explosively along with the growth of voxel resolution, most current state-of-the-arts have to tailor their framework into a low-resolution representation with the sacrifice of detail prediction. Thus, voxel resolution becomes one of the crucial difficulties that lead to the performance bottleneck.

In this paper, we propose to devise a new geometry-based strategy to embed depth information with low-resolution voxel representation, which could still be able to encode sufficient geometric information, $e.g.$, room layout, object's sizes and shapes, to infer the invisible areas of the scene with well structure-preserving details. To this end, we first propose a novel 3D sketch-aware feature embedding to explicitly encode geometric information effectively and efficiently. With the 3D sketch in hand, we further devise a simple yet effective semantic scene completion framework that incorporates a light-weight 3D Sketch Hallucination module to guide the inference of occupancy and the semantic labels via a semi-supervised structure prior learning strategy. We demonstrate that our proposed geometric embedding works better than the depth feature learning from habitual SSC frameworks. Our final model surpasses state-of-the-arts consistently on three public benchmarks, which only requires 3D volumes of \hspace{0.0cm} $60\times 36\times 60$ resolution for both input and output. The code and the supplementary material will be available at \href{https://charlesCXK.github.io}{https://charlesCXK.github.io}.
\end{abstract}

\section{Introduction}
Semantic Scene Completion (SSC), which provides an alternative to understand the 3D world with both 3D geometry and semantics of the scene from a partial observation, is an emerging topic in computer vision for its wide applicability on many applications, \eg, augmented reality, surveillance and robotics. Due to the high memory and computational cost requirements on inherent voxel representation, most existing methods~\cite{song2017semantic-sscnet,guo2018view-vvnet,zhang2019cascaded-ccpnet,garbade2018two-ts3d,li2019rgbd-ddrnet,liu2018see-satnet,zhang2018semantic,dourado2019edgenet} achieve semantic scene completion through sophisticated 3D context modeling on \textit{implicitly embedded depth feature} that learnt from general 3D CNNs. These methods are either error-prone on classifying fine details of objects or have the difficulties in completing the scene when there exists a large portion of geometry missing, as shown in Figure \ref{fig:intro}.

Several recent studies~\cite{garbade2018two-ts3d,li2019rgbd-ddrnet,liu2018see-satnet} present promising results on this topic by introducing high-resolution RGB images into the process. Though driven by various motivations, these methods could be thought as building \textit{cross-modality feature embedding}  with the assumption that the fine detail feature could be compensated from RGB counterpart and computation-efficient property could be guaranteed with 2D operators on RGB source. However, such an approach is highly relied on the effectiveness of cross-modality feature embedding module design and is vulnerable to complex scenes.

In contrast, from the human perception, it is a breeze to complete and recognize 3D scene even from the partial low-resolution observation, due to the prior knowledge on object's geometry properties, \eg, size and shape, of different categories. From this perspective, we hypothesize the feature embedding strategy that explicitly encodes the geometric information could facilitate the network learning the concept of object's structure, and therefore reconstructing and recognizing the scene precisely even from the low-resolution partial observation. To this end, the geometry properties need to be resolution-invariant or at least resolution-insensitive.

\begin{figure}[ht]
\centering
\includegraphics[width=\columnwidth]{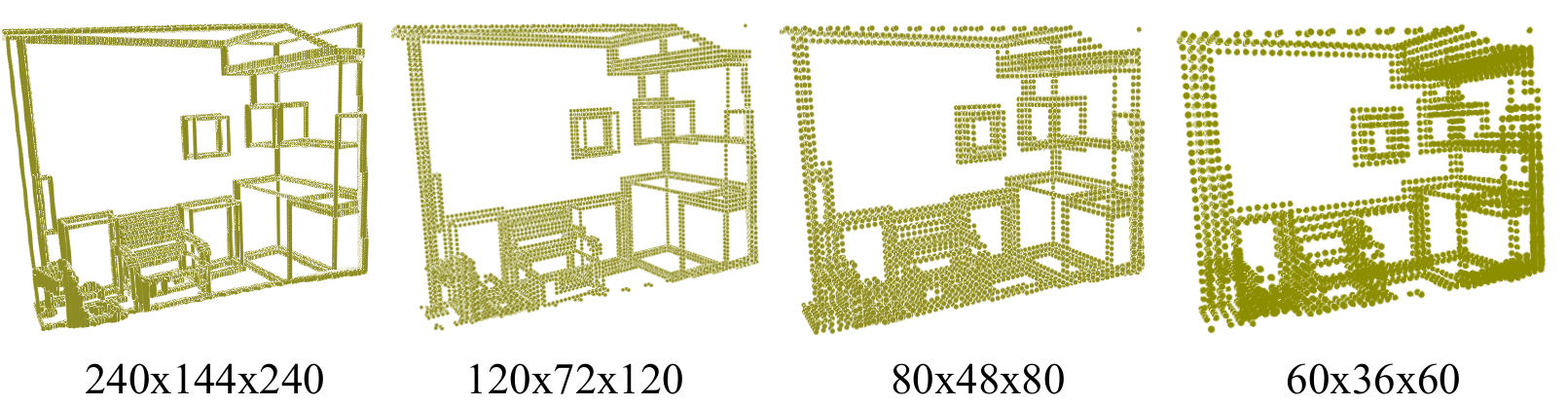}
\caption{\textbf{Visualization of sketches extracted from semantic labels with different resolutions.} From left to right, the sketch begins to lose some details as resolution decreases, while the structure description of the scene is well preserved.}
\label{fig:sketch}
\end{figure}

Based on this intuition, we present 3D sketch\footnote{3D Sketch could be understood as a kind of 3D boundary. To distinguish it with the concept of \textit{edge}/\textit{boundary} in image space, we refer it as \textit{3D Sketch}. }-aware feature embedding, an explicit and compact depth feature embedding schema for the semantic scene completion task. It has been demonstrated in ~\cite{shi2017structure} that the similar geometric cue in image space, \ie, 2D boundary, is resolution-insensitive. We show that the 3D world also holds the same conclusion, as indicated in Figure \ref{fig:sketch}.

However, 3D sketch extracted from 2D depth image is still a 2D/2.5D observation from a single viewpoint. To fully utilize the strength of this new feature embedding, we further propose a 3D sketch-aware semantic scene completion network, which injects a 3D Sketch Hallucination Module to infer the full 3D sketch from the partial one at first, and then utilize the feature embedded from the hallucinated 3D sketch to guide the reconstruction and recognition. Specifically, since lifting the 2D/2.5D observation to full 3D sketch is intrinsically ambiguous, instead of directly regressing the ground-truth full 3D sketch, we seek a nature prior distribution to sample diverse reasonable 3D sketches. We achieve that by tailoring Conditional Variational Autoencoder (CVAE)~\cite{sohn2015learning} into the
3D Sketch Hallucination Module design. We show that such a design could help to generate accurate and realistic results even when there is a large portion of geometry missing from the partial observation.

We summarize our contributions as follows: 
\begin{itemize}
    \item We devise a new geometric embedding from depth information, namely 3D sketch-aware feature embedding, to break the performance bottleneck of the SSC task caused by a low-resolution voxel representation.
    \item We introduce a simple yet effective semantic scene completion framework that incorporates a novel 3D Sketch Hallucination Module to guide the full 3D sketch inference from partial observation via semi-supervised structure prior property of Conditional Variational Autoencoder (CVAE), and utilizes the feature embedded from the hallucinated 3D sketch to further guide the scene completion and semantic segmentation.
    \item Our model outperforms state-of-the-arts consistently on three public benchmarks, with only requiring 3D volumes of $60 \times 36 \times 60$ resolution for both input and output. 

\end{itemize}

\section{Related Work}
\subsection{Object Shape Completion}
Object shape completion has a long history in geometry processing. We summarize existing methods to two categories: knowledge-based and learning-based.

Knowledge-based methods complete partial input of an object by reasoning geometric cues or matching it with 3D models from an extensive shape database. Some works detect symmetries in meshes or point clouds and use them to fill in missing data, such as~\cite{thrun2005shape,speciale2016symmetry,pauly2008discovering}.
An alternative is to match the partial input with CAD models from a large database~\cite{nan2012search,shao2012interactive,kim2012acquiring}. However, it is too expensive to retrieval, and it has poor generalization for new shapes that do not exist in the database.

Learning-based methods are more flexible and effective than knowledge-based ones. They usually infer the invisible area with a deep neural network, which has fast inference speed and better robustness.~\cite{dai2017shape} proposes a 3D-Encoder-Predictor Network, which first encodes the known and unknown space to get a relatively low-resolution prediction, and then correlates this intermediary result with 3D geometry from a shape database.~\cite{yuan2018pcn} proposes an end-to-end method that directly operates on raw point clouds without any structural assumption about the underlying shape.~\cite{stutz2018learning} proposes a weakly-supervised approach that learns a shape prior on synthetic data and then conducts maximum likelihood fitting using deep neural networks.

These methods focus on reconstructing 3D shape from the partial input of a single object, which makes it hard for them to extend to partial scenes along with multiple objects estimated in semantic level.

\subsection{Semantic Scene Completion}
Semantic Scene Completion (SSC) is a fundamental task in 3D scene understanding, which produces a complete 3D voxel representation of volumetric occupancy and semantic labels. SSCNet~\cite{song2017semantic-sscnet} is the first to combine these two tasks in an end-to-end way. ESSCNet~\cite{zhang2018efficient-esscnet} introduces Spatial Group Convolution (SGC) that divides input volume into different groups and conduct 3D sparse convolution on them. VVNet~\cite{guo2018view-vvnet} combines 2D and 3D CNN with a differentiable projection layer to efficiently reduce computational cost and enable feature extraction from multi-channel inputs. ForkNet~\cite{wang2019forknet} proposes a multi-branch architecture and draws on the idea of generative models to sample new pairs of training data, which alleviates the limited training samples problem on real scenes. CCPNet~\cite{zhang2019cascaded-ccpnet} proposes a self-cascaded context aggregation method to reduce semantic gaps of multi-scale 3D contexts and incorporates local geometric details in a coarse-to-fine manner. 

Some works also utilize RGB images as vital complementary to depth. TS3D~\cite{garbade2018two-ts3d} designs a two-stream approach to leverage semantic and depth information, fused by a vanilla 3DCNN. SATNet~\cite{liu2018see-satnet} disentangles semantic scene completion task by sequentially accomplishing 2D semantic segmentation and 3D semantic scene completion tasks. DDRNet~\cite{li2019rgbd-ddrnet} proposes a light-weight Dimensional Decomposition Residual network and fused multi-scale RGB-D features seamlessly.

 Above methods could be regraded as encoding depth information implicitly by either single- or cross-modality feature embedding. They map depth information into an inexplicable high-dimensional feature space and then use the feature to predict the result directly. Different from current methods, we propose an explicit geometric embedding strategy from depth information, which predicts 3D sketch first and utilize the feature embedded from it to guide the reconstruction and recognition.

\subsection{2D Boundary Detection}
2D Boundary detection is a fundamental challenge in computer vision. There are lots of methods proposed to detect boundaries. Sobel operator~\cite{sobel1968operator} and Canny operator~\cite{canny1986computational} are two hand-craft based classics that detect boundaries with gradients of the image. Learning-based works ~\cite{liu2017richer,he2019bi,yang2016object} try to employ deep neural networks with supervision. Most of them directly concatenate multi-level features to extract the boundary. Since boundary includes a distinct geometric structure of objects, some other works try to inject boundary detection into other tasks to help boost the performance. ~\cite{wang2019salient} combines boundary detection with salient object detection task to encourage better edge-preserving salient object segmentation. \cite{yu2018learning,takikawa2019gated} introduce boundary detection into semantic segmentation task to obtain more precise semantic segmentation results. ~\cite{wayne2018lab} achieves robust facial landmark detection by utilizing facial boundary as an intermediate representation to remove the ambiguities. With similar spirits, we introduce a 3D sketch-aware feature embedding to break the performance bottleneck of the SSC task caused by a low-resolution voxel representation.

\subsection{Structure Representation Learning}
Deep generative models have demonstrated significant performance in structure representation learning.~\cite{sohn2015learning} develops a deep conditional generative model to predict structured output using Gaussian latent variables, which can be trained efficiently in the framework of stochastic gradient variational Bayes.~\cite{zhang2018unsupervised} proposes an autoencoding formulation to discover landmarks as explicit structural representations in an unsupervised manner.~\cite{esser2018variational} proposes to synthesize images under the guidance of shape representations and conditions on the learned textural information.~\cite{sharma2019monocular} employs CVAE to stress the issue of the inherent ambiguity in 2D-to-3D lifting in the pose estimation task.
Adopting the idea of structure representation learning, we embed the geometric structure of a 3D scene through a CVAE~\cite{sohn2015learning} conditioned on the estimated sketch.

\begin{figure*}[t!]
\centering
\includegraphics[width=\textwidth]{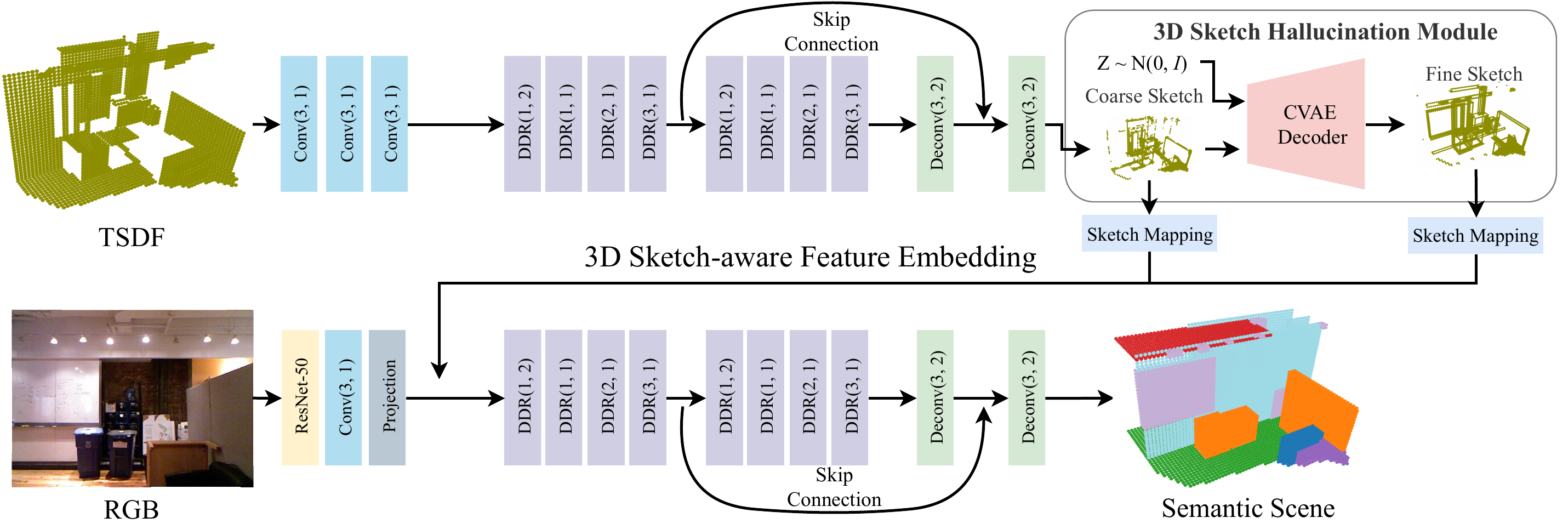}
\caption{\small{\textbf{Overview of our network}. We first generate structure prior information from the TSDF input and use CVAE to refine the prediction. Then the prior information will be passed to the RGB-branch to predict occupancy and object labels for each voxel in the view frustum. The convolution parameters are shown as (kernel size, dilation). The DDR parameters are shown as (dilation, downsample rate). The Deconvolution parameters are shown as (kernel size, upsample rate).}}
\vspace{-2mm}
\label{fig:arch}
\end{figure*}

\section{Methodology}

The overall architecture of our network is illustrated in Figure \ref{fig:arch}. The proposed method consists of multiple stages and each stage adopts an encoder-decoder architecture. Taking a pair of RGB and depth images of a 3D scene as input, the network outputs a  dense prediction and each voxel in the view frustum is assigned with a semantic label $C_i$, where $i \in [0,1, \cdots,N]$ and $N$ is the number of semantic categories. $C_0$ stands for empty voxels.

More specifically, we stack two stages and let each stage handle different tasks. The first stage tackles the task of sketch extraction. It embeds the geometric cues contained in the scene and provides the structure prior information (which we call it sketch) for the next stage. Besides, we employ CVAE to guide the predicted sketch. The second stage tackles the task of semantic scene completion (SSC) based on the extracted sketch. Details are introduced below.

\subsection{Generation of Ground-truth Sketch} We perform 3D Sobel operator on the semantic label to extract the sketch of the semantic scene. Suppose we have obtained gradients $g_x^i, g_y^i, g_z^i$ at the $i$-th voxel $V_i$ along $x,y,z$ axes, we first binarize these values to be $0$ or $1$ to eliminate the semantic gap. For example, the gap between class 1 and class 2 should be considered equal to the gap between class 1 and class 10 when generating the sketch. Finally, the extracted sketch can be described as a set: $\mathbf{S_{sketch}} = \{V_i: g_x^i + g_y^i + g_z^i > 1\}$. To distinguish generated geometric representation with generally 2D edge/boundary, we refer it as \textit{3D Sketch}.

\subsection{Sketch Prediction Stage}
This stage takes a single-view depth map as input and encodes it as a 3D volume. We follow~\cite{song2017semantic-sscnet} to rotate the scene to align with gravity and room orientation based on Manhattan assumption. We adopt Truncated Signed Distance Function (TSDF) to encode the 3D space, where every voxel stores the distance value $d$ to its closest surface and the sign of the value indicates whether the voxel is in free space or occluded space. The encoder volume has a grid size of $0.02$~m and a truncation value of $0.24$~m, resulting in a $240 \times 144 \times 240$ volume. For the saving of computational cost,~\cite{song2017semantic-sscnet} downsamples the ground truth by a rate of $4$, and we use the same setting. Following SATNet~\cite{liu2018see-satnet}, we also downsample the input volume by a rate of $4$ and use $60 \times 36 \times 60$ resolution as input.

Previous works~\cite{peng2017large,zhang2018context, yu2018learning} demonstrate that contextual information is important for 2D semantic segmentation. Due to the sparseness and the high computational cost of 3D voxels, it is hard to obtain the context of the scene. To learn rich contextual information, we should make sure that our network has a large enough receptive field without significantly increasing the computational cost. To this end,~\cite{li2019rgbd-ddrnet} proposed Dimensional Decomposition Residual (DDR) block which is computation-efficient compared with basic 3D residual block. We adopt DDR block as our basic unit and stack them layer by layer with different dilation rates to maintain big receptive fields. As shown in Figure \ref{fig:arch}, We first employ several convolutions to encode the TSDF volume into high dimensional features.  Then we aggregate the contextual information of the input feature by several DDR blocks and downsample it by a rate of $4$ to reduce computational cost. Finally, we employ two deconvolution layers to upsample the feature volume and obtain the dense predicted sketch, which we denote as $\hat G_{raw}$. Following~\cite{song2017semantic-sscnet}, we add a skip connection between two layers for better gradient propagation, which is illustrated in Figure \ref{fig:arch}.

Due to the input of semantic scene completion task is not a complete scene, we assume that a more precise and complete sketch will bring more information increments to the subsequent stage. To some extent, it may make up for the inadequacy of incomplete input. Thus we design a 3D Sketch Hallucination Module to handle this issue.

\subsection{3D Sketch Hallucination Module}
Lifting 2D/2.5D observation to full 3D sketch is intrinsically ambiguous, we thus seek a nature prior distribution to sample diverse reasonable 3D sketches instead of directly regressing the ground truth. Thus, we employ CVAE to further process the original predicted sketch by sampling an accurate and diverse sketch set $S = \{\hat G^k_{refined} : k \in {1, 2, ..., K} \}$ conditioned on the estimated $\hat G_{raw}$.

\begin{figure}[ht]
\centering
\includegraphics[width=\columnwidth]{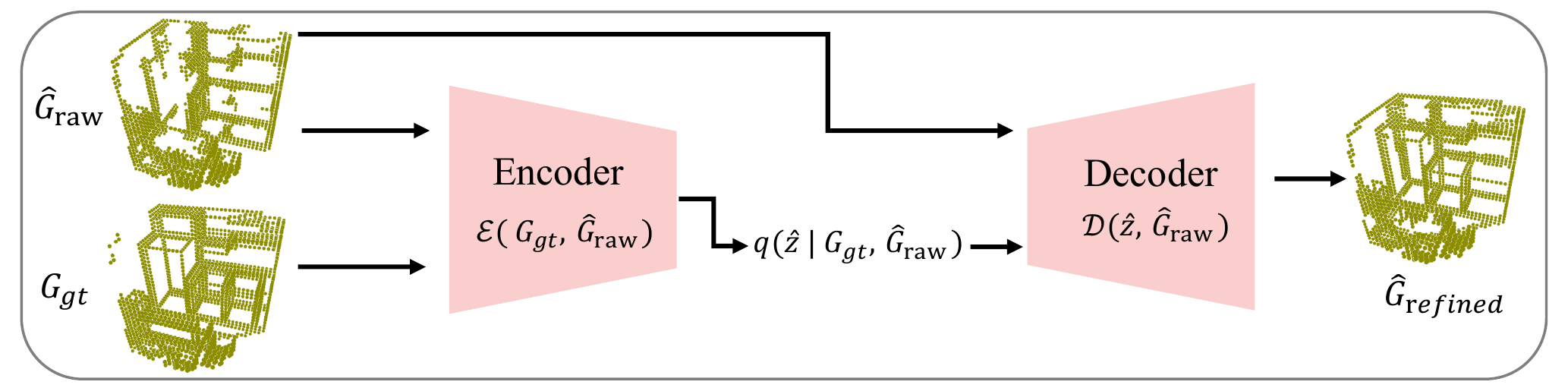}
\caption{\small{\textbf{Architecture of the proposed Sketch Hallucination Module}. During training time, the original estimated sketch and the ground-truth sketch are fed into the encoder to generate mean and diagonal covariance for the posterior $q$. Then the decoder will reconstruct the ground-truth sketch with a latent sampled from $q$ and the original estimated sketch as input.}}
\vspace{-2mm}
\label{fig:arch-cvae}
\end{figure}

The proposed 3D Sketch Hallucination Module (as shown in Figure \ref{fig:arch-cvae}) consists of a standard encoder-decoder structure. The encoder which we denote as $\mathcal{E}(G_{gt}, \hat G_{raw})$, performs some convolution operations on the input ground-truth sketch and a condition $\hat G_{raw}$ to output the mean and diagonal covariance for the posterior $q(\hat{z} | G_{gt}, \hat G_{raw})$. Then the decoder which we denote as $\mathcal{D}(\hat{z}, \hat G_{raw})$ will reconstruct the sketch by taking a latent $\hat z$ sampled from the posterior $q(\hat{z} | G_{gt}, \hat G_{raw})$ and the condition $\hat G_{raw}$ as input.

During training, we optimize the proposed module though minimizing the following objective function,
\begin{equation}
    \begin{aligned}
    \mathcal{L}_{\rm CVAE} &=
    \lambda_1 KL(q(\hat{z} | G_{gt}, \hat G_{raw}) ~||~ p(z | \hat G_{raw}))  \\
    & + \lambda_2 \mathbb{E}_{z \sim q(\hat{z} | G_{gt}, \hat G_{raw})} \boldsymbol{\epsilon}(G_{gt}, \mathcal{D}(\hat{z}, \hat G_{raw})), \label{eq:cvae} \\
    \end{aligned}
\end{equation}
where $\boldsymbol{\epsilon}$ is a cross-entropy loss and $KL(x || y)$ is the Kullback-Leibler divergence loss.  We use $\lambda_i$ as hyper-parameter to weight these two loss items. $\mathbb{E}$ is the expectation which is taken over $K$ samples. The $p(z |\hat G_{raw})$ is the prior distribution. To ensure gradients can be backpropagated through the latent code, the KL divergence is required to be computed in a closed form. Thus, the latent space of CVAE is typically restricted to be a distribution over $\mathcal{N} (0, I)$. We follow this setting in our framework. Specifically, it draws a Gaussian prior assumption over the coarse-step geometry representation to fine-step geometry representation in our framework. Sketch is a simple yet compact geometry representation which suits the assumption. Since the encoder will not be used during inference, the current objective will introduce inconsistency between training and inference. To address this issue, we follow~\cite{sohn2015learning,sharma2019monocular} to set the encoder the same as prior network $p(z) \sim \mathcal{N} (0, I)$, namely $Gaussian \ Stochastic \  Neural \  Network$ (GSNN) and the reparameterization trick of CVAE can be used to train GSNN. We combine $\mathcal{L}_{\rm GSNN}$ and $\mathcal{L}_{\rm CVAE}$ with $\alpha$ as weight term to obtain final objective for our refine network,
\begin{align}
    \mathcal{L}_{\rm GSNN} & = \mathbb{E}_{z \sim \mathcal{N}(0, I)} \boldsymbol{\epsilon}(G_{gt}, \mathcal{D}(z, \hat G_{raw})), \\
    \mathcal{L}_{\rm hybrid} & = \mathcal{L}_{\rm CVAE} + \alpha \mathcal{L}_{\rm GSNN}, \label{eq:hybrid}
\end{align}

Durning inference, we randomly sample $z$ from $\mathcal{N}(0, I)$ for $K$ times and obtain $K$ different $\mathcal{D}(z, \hat G_{raw})$, which are denoted as $S = \{\hat G^k_{refined} : k \in {1, 2, ..., K} \}$. We average them and obtain the refined sketch $\hat G_{refined}$.

\subsection{Semantic Scene Completion Stage}

In this stage, we will take a single RGB image and the pre-computed sketches from the former stage as input to densely predict the semantic scene labels. We divide this stage into three parts: 2D feature learning, 2D-3D projection and 3D feature learning. The input RGB image is firstly fed into a ResNet-50~\cite{he2016deep} to extract local and global textural features. For achieving stable training, we utilize the parameters pre-trained on ImageNet~\cite{deng2009imagenet} and freeze the weight of them. Due to the output tensor of ResNet-50 has too many channels, which will bring too much computational cost for 3D learning part, we adopt a convolution layer followed by a Batch Normalization~\cite{ioffe2015batch} and Rectified Linear Unit (ReLU) to reduce its dimensions.

Then the computed 2D semantic feature map will be projected into 3D space according to the depth map and the corresponding camera parameters. Given the depth image $I_{depth}$, the intrinsic camera matrix $K_{camera} \in \mathbb{R}^{3 \times 3}$, and the extrinsic camera matrix $E_{camera} \in \mathbb{R}^{3 \times 4}$, each pixel $p_{u, v}$ in the 2D feature map can be projected to an individual 3D point $p_{x, y, z}$. Because the resolution of the 3D volume is lower than the 2D feature map, multiple points may be divided into the same voxel in the process of voxelization. For those voxels, we only keep one feature vector in a certain voxel by max-pooling. After this step, the semantic feature vector for each pixel is assigned to its corresponding voxel via the mapping $\mathbb{M}$. Since many areas are not visible, zero vectors are assigned to the occluded areas and empty foreground in the scene.

Given the projected 3D feature map $\mathbf{F}_{proj} \in  \mathbb{R}^{C \times H \times W \times L} $, where $C$ is the number of channels and $H,W,L$ are size of the feature map. We now use the prior information $\hat G_{raw}$ and $\hat G_{refined}$ as guidance. We define two sketch mappings: $\mathcal{F}_{raw}: \hat G_{raw} \rightarrow \mathbf{F}_{raw} \in \mathbb{R}^{C \times H \times W \times L} $ and $\mathcal{F}_{refined}: \hat G_{refined} \rightarrow \mathbf{F}_{refined} \in \mathbb{R}^{C \times H \times W \times L} $ to map these prior information to the same feature space with $\mathbf{F}_{proj}$. After these two mapping operations, both $\mathbf{F}_{raw}$ and $\mathbf{F}_{refined}$ have the same resolution and dimension with $\mathbf{F}_{proj}$. Thus we introduce the prior information by an element-wise addition operation on $\mathbf{F}_{proj}$, $\mathbf{F}_{raw}$ and $\mathbf{F}_{refined}$. In pratice, these two mapping functions are implemented by $3 \times 3$ convolution layers. In the following, the new feature map will be fed into a 3D CNN, whose architecture is the same with that of sketch-branch, and we obtain the final semantic scene completion predictions.

\subsection{Loss Function}
During training, the dataset is organized as a set $\{(X_{\rm TSDF}, X_{\rm RGB}, G_{gt}, S_{gt})\}$, where $G_{gt}$ represents the ground-truth sketch and $S_{gt}$ represents the ground-truth semantic labels. We optimize the entire architecture by the following formulas:
\begin{align}
    & \mathcal{L}_{\rm loss} = \mathcal{L}_{\rm semantic} +  \mathcal{L}_{\rm hybrid} + \mathcal{L}_{\rm sketch}, \\
    & \mathcal{L}_{\rm semantic} = \boldsymbol{\epsilon}(S_{gt}, \mathcal{D}_s (\mathcal{E}_s (X_{\rm RGB}))), \\
    & \mathcal{L}_{\rm sketch} = \boldsymbol{\epsilon}(G_{gt}, \mathcal{D}_g (\mathcal{E}_g (X_{\rm TSDF})),
\end{align}
where $\mathcal{D}_g, \mathcal{E}_g$ are the encoder and the decoder of the sketch stage, $\mathcal{D}_s, \mathcal{E}_s$ are the encoder and the decoder of the semantic stage, $\mathcal{L}_{\rm hybrid}$ is defined in Eq.~\eqref{eq:hybrid}, and $\boldsymbol{\epsilon}$ denotes the cross-entropy loss.


\section{Experiments}
\subsection{Datasets and Evaluation Metrics}
We evaluate the proposed method on three datasets: NYU Depth V2~\cite{nyudv2} (which is denoted as NYU in the following), NYUCAD~\cite{firman2016structured} and SUNCG~\cite{song2017semantic-sscnet}. We will introduce these three datasets in detail in the supplementary material. We follow SSCNet~\cite{song2017semantic-sscnet} and use precision, recall and voxel-level intersection over union (IoU) as evaluation metrics. Following~\cite{song2017semantic-sscnet}, two tasks are considered: semantic scene completion (SSC) and scene completion (SC). For the task of SSC, we evaluate the IoU of each object class on both observed and occluded voxels in the view frustum. For the task of SC, we treat all voxels as binary predictions, $i.e.$, empty or non-empty. We evaluate the binary IoU on occluded voxels in the view frustum.

\subsection{Implementation Details}

\noindent \textbf{Training Details.} We use PyTorch framwork to implement our experiments with $2$ GeForce GTX 1080 Ti GPUs. We adopt mini-batch SGD with momentum to train our model with batch size $4$, momentum $0.9$ and weight decay $0.0005$. We employ a poly learning rate policy where the initial learning rate is multiplied by $(1-\frac{iter}{max\_iter})^{0.9}$. For both NYU and NYUCAD, we train our network for $250$ epochs with initial learning rate $0.1$. For SUNCG, we train our network for $8$ epochs with initial learning rate $0.01$. The expection in Eq.~\eqref{eq:cvae} is estimated using $K$ = 4 samples. $\lambda_1$, $\lambda_2$ and $\alpha$ in Eq.~\eqref{eq:cvae} and Eq.~\eqref{eq:hybrid} are set to $2$, $1$ and $1.5$ respectively.

\begin{table}[htbp]
\begin{center}
\resizebox{0.7\columnwidth}{!}{
\begin{tabular}{|c | c | c |} 
\hline
\textbf{Drop Rate(\%)} & \textbf{SC-IoU(\%)} & \textbf{SSC-mIoU(\%)} \\
\hline\hline
0 & \textbf{94.2} & \textbf{65.0} \\
20 & 93.7 & 63.6 \\
40 & 93.2 & 62.3 \\
60 & 92.0 & 59.9 \\
80 & 89.9 & 57.1 \\
\hline
\end{tabular}
}
\vspace{-0.2cm}
\end{center}
\caption{\textbf{Oracle Ablation}. \textit{(Oracle) Drop Rate} means we randomly drop the ground-truth sketch in a certain proportion. We perform this ablation study on NYUCAD dataset.}
\label{tab:oracle}
\end{table}

\noindent \textbf{Oracle Ablation.} To obtain the theoretical upper limit of the proposed method, we replace the output of the first stage with the ground-truth 3D sketch to supply the structure prior. Results are shown in Table \ref{tab:oracle}. $Drop\ Rate$ means we randomly discard some voxels in the ground-truth 3D sketch by some ratio. We observe that with the whole 3D sketch as structure prior, our network could infer most of the invisible areas and obtain $94.2\%$ SC IoU. As the drop rate increases to $80\%$, the performance has not dropped a lot and is still higher than the best performance of the proposed method, which verifies the validity of accurate structure prior.

\subsection{Comparisons with State-of-the-art Methods}
We further compare the proposed method with state-of-the-art methods. Table \ref{tab:SotaOnNYU} shows the performances by state-of-the-art methods on NYU dataset. We observe that the proposed method outperforms all existing methods by a large margin, more specifically, we gain an increase of $7.8\%$ SC IoU and $2.6\%$ SSC mIoU compared to CCPNet \cite{zhang2019cascaded-ccpnet}. We argue that this improvement is caused by the novel two-stage architecture which makes the full use of the structure prior. The provided structure prior can accurately infer invisible areas of the scene with well structure-preserving details.

We also conduct experiments on NYUCAD dataset to validate the generalization of the proposed method. Table \ref{tab:SotaOnNYUCAD} presents the quantitative results on NYUCAD dataset. Our proposed method maintains the performance advantage and outperforms CCPNet \cite{zhang2019cascaded-ccpnet} by $1.8\%$ SC IoU and $2.0\%$ SSC mIoU. Note that although some works \cite{zhang2019cascaded-ccpnet,wang2019forknet,garbade2018two-ts3d} use larger input resolution than ours, the proposed method still outperforms them with a low-resolution input of $60 \times 36 \times 60$.

Experiments on SUNCG dataset and the visualization of the SSC results compared with SSCNet~\cite{song2017semantic-sscnet} on NYUCAD dataset are put in the supplementary material.

\subsection{Ablation Study}
To evaluate the effectiveness of the pivotal components of our method, we perform extensive ablation studies using the same hyperparameters. Details are illustrated below.

\begin{table}[ht]
\begin{center}
\resizebox{\columnwidth}{!}{
\begin{tabular}{|c|c|c|c|c|} 
\hline
\textbf{\#Stage} & \textbf{Structure Prior} &  \textbf{CVAE} & \textbf{SC-IoU(\%)} & \textbf{SSC-mIoU(\%)} \\
\hline\hline
1 & \xmark & \xmark  & 79.3 & 48.7\\
2 & \xmark & \xmark  & 81.1 & 50.6\\
2 & \cmark & \xmark  & 83.6 & 53.9\\
2 & \cmark & \cmark  & \textbf{84.2} & \textbf{55.2}\\
\hline
\end{tabular}
}
\vspace{-0.35cm}
\end{center}
\caption{\textbf{Ablation studies on different modules}. We perform this ablation study on NYUCAD dataset.}
\label{tab:ablation-nyucad}
\end{table}

\noindent \textbf{Different Modules in the Framework.}
We first conduct ablation studies on different modules in the proposed method. Results are shown in Table \ref{tab:ablation-nyucad}. From Row 1 and Row 2, we find that just adopting a dual-path structure could boost the performance, as more parameters are introduced. In the third row, with the introduction of structure prior, our network could infer the invisible areas of the scene with well structure-preserving details, which brings great improvements. Finally, with the proposed 3D Sketch Hallucination Module, we further boost the performance and achieve $84.2\%$ SC IoU and $55.2\%$ SSC mIoU, which are both new state-of-the-art performance on NYUCAD.

\begin{table*}[ht]
\begin{center}
\resizebox{\textwidth}{!}{
\begin{tabular}{|l| c | c |c c c|c c c c c c c c c c c | c|} 
\hline
  & & & \multicolumn{3}{c|}{scene completion} & \multicolumn{12}{c|}{semantic scene completion} \\ \hline
Methods  & Resolution & Trained on & prec. & recall & IoU & ceil. & floor & wall & win. & chair & bed & sofa & table & tvs & furn. & objs. & avg. \\ 
\hline
Lin \etal~\cite{lin2013holistic}  & $(240, 60)$ & NYU & 58.5 & 49.9 & 36.4 &  0.0 & 11.7 & 13.3 & 14.1 &  9.4 & 29.0 & 24.0 &  6.0 &  7.0 & 16.2 &  1.1 & 12.0\\
Geiger \etal~\cite{geiger2015joint} & $(240, 60)$ & NYU & 65.7 & 58.0 & 44.4 & 10.2 & 62.5 & 19.1 &  5.8 &  8.5 & 40.6 & 27.7 &  7.0 &  6.0 & 22.6 &  5.9 & 19.6\\ 
SSCNet~\cite{song2017semantic-sscnet} & $(240, 60)$ & NYU & 57.0 & \textbf{94.5} & 55.1 & 15.1 & 94.7 & 24.4 &  0.0 & 12.6 & 32.1 & 35.0 & 13.0 &  7.8 & 27.1 & 10.1 & 24.7\\
ESSCNet~\cite{zhang2018efficient-esscnet} & $(240, 60)$ & NYU & 71.9 & 71.9 & 56.2 & 17.5 & 75.4 & 25.8 &  6.7 & 15.3 & 53.8 & 42.4 & 11.2 &    0 & 33.4 & 11.8 & 26.7\\ 
DDRNet~\cite{li2019rgbd-ddrnet}* & $(240, 60)$ & NYU  & 71.5  & 80.8 & 61.0 & 21.1 & 92.2 & 33.5 & 6.8 & 14.8 & 48.3 & 42.3 & 13.2 & 13.9 & 35.3 &  13.2 & 30.4\\ 
VVNetR-120~\cite{guo2018view-vvnet}  & $(120, 60)$ & NYU+SUNCG & 69.8 & 83.1 & 61.1 & 19.3 & 94.8 & 28.0 & 12.2 & 19.6 & 57.0 & 50.5 & 17.6 & 11.9 & 35.6 & 15.3 & 32.9  \\
TS3D~\cite{garbade2018two-ts3d}* & $(240, 60)$ & NYU  & - & - & 60.0 & 9.7 & 93.4 & 25.5 & 21.0 & 17.4 & 55.9 & 49.2 & 17.0 & 27.5 & 39.4 & 19.3 & 34.1\\
SATNet-TNetFuse~\cite{liu2018see-satnet}* & $(60, 60)$ & NYU+SUNCG & 67.3 & 85.8 & 60.6 & 17.3 & 92.1 & 28.0 & 16.6 & 19.3 & 57.5 & 53.8 & 17.2 & 18.5 & 38.4 & 18.9 & 34.4 \\
ForkNet~\cite{wang2019forknet}  & $(80, 80)$ & NYU & - & - & 63.4 & 36.2 & 93.8 & 29.2 & 18.9 & 17.7 & \textbf{61.6} & 52.9 & 23.3 & 19.5 & \textbf{45.4} & 20.0 & 37.1 \\
CCPNet~\cite{zhang2019cascaded-ccpnet} & $(240, 240)$ & NYU  & 74.2  & 90.8 & 63.5 & 23.5 & \textbf{96.3} & 35.7 & 20.2 & 25.8 & 61.4 & \textbf{56.1} & 18.1 & \textbf{28.1} & 37.8 & 20.1 & 38.5\\ 
\hline
Ours* & $(60, 60)$ & NYU  & \textbf{85.0} & 81.6 & \textbf{71.3} & \textbf{43.1} & 93.6 & \textbf{40.5} & \textbf{24.3} & \textbf{30.0} & 57.1 & 49.3 & \textbf{29.2} & 14.3 & 42.5 & \textbf{28.6} & \textbf{41.1} \\ 
\hline
\end{tabular}
}
\vspace{-0.35cm}
\end{center}
\caption{\textbf{Results on NYU dataset}. Bold numbers represent the best scores. \textit{Resolution(a, b)} means the input resolution is $(a \times 0.6a \times a)$ and the output resolution is $(b \times 0.6b \times b)$. `*' are RGB-D based methods.}
\label{tab:SotaOnNYU}
\end{table*}

\begin{table*}[ht]
\begin{center}
\resizebox{\textwidth}{!}{
\begin{tabular}{|l| c | c |c c c|c c c c c c c c c c c | c|} 
\hline
  & & & \multicolumn{3}{c|}{scene completion} & \multicolumn{12}{c|}{semantic scene completion} \\ \hline
Methods  & Resolution & Trained on & prec. & recall & IoU & ceil. & floor & wall & win. & chair & bed & sofa & table & tvs & furn. & objs. & avg. \\ 
\hline
Zheng \etal~\cite{zheng2013beyond} & $(240, 60)$ & NYUCAD	& 60.1 & 46.7 & 34.6 & - & - & - & - & - & - & - & - & - & - & - & - \\ 
Firman \etal~\cite{firman2016structured} & $(240, 60)$ & NYUCAD	& 66.5 & 69.7 & 50.8 & - & - & - & - & - & - & - & - & - & - & - & - \\ 
SSCNet~\cite{song2017semantic-sscnet} & $(240, 60)$ & NYUCAD+SUNCG  & 75.4 & \textbf{96.3} & 73.2 & 32.5 & 92.6 & 40.2 &  8.9 & 33.9 & 57.0 & 59.5 & 28.3 &  8.1 & 44.8 & 25.1 & 40.0\\ 
VVNetR-120~\cite{guo2018view-vvnet} & $(120, 60)$ & NYUCAD+SUNCG & 86.4 & 92.0 & 80.3 & - & - & - & - & - & - & - & - & - & - & - & - \\
DDRNet~\cite{li2019rgbd-ddrnet}* & $(240, 60)$	& NYUCAD & 88.7 & 88.5 & 79.4 & 54.1 & 91.5 & 56.4 & 14.9 & 37.0 & 55.7 & 51.0 & 28.8 & 9.2 & 44.1 & 27.8 & 42.8 \\
TS3D~\cite{garbade2018two-ts3d}* & $(240, 60)$   & NYUCAD & - & - & 76.1 & 25.9 & 93.8 & 48.9 & 33.4 & 31.2 & 66.1 & 56.4 & 31.6 & \textbf{38.5} & 51.4 & 30.8 & 46.2 \\ 
CCPNet~\cite{zhang2019cascaded-ccpnet} & $(240, 240)$ & NYUCAD & \textbf{91.3} & 92.6 & 82.4 & 56.2 & \textbf{94.6} & 58.7 & \textbf{35.1} & 44.8 & 68.6 & 65.3 & 37.6 & 35.5 & 53.1 & 35.2 & 53.2 \\
\hline
Ours* & $(60, 60)$ & NYUCAD  & 90.6 & 92.2 & \textbf{84.2} & \textbf{59.7} & 94.3 & \textbf{64.3} & 32.6 & \textbf{51.7} & \textbf{72.0} & \textbf{68.7} & \textbf{45.9} & 19.0 & \textbf{60.5} & \textbf{38.5} &  \textbf{55.2} \\ 
\hline
\end{tabular}
}
\vspace{-0.35cm}
\end{center}
\caption{\textbf{Results on NYUCAD dataset}. Bold numbers represent the best scores. \textit{Resolution(a, b)} means the input resolution is $(a \times 0.6a \times a)$ and the output resolution is $(b \times 0.6b \times b)$. `*' are RGB-D based methods.}
\label{tab:SotaOnNYUCAD}
\end{table*}

\begin{table}[t]
\begin{center}
\resizebox{\columnwidth}{!}{
\begin{tabular}{|c|c|c|c|c|c|} 
\hline
\textbf{Input} & \textbf{Shape} &  \textbf{Semantic Labels} & \textbf{Sketch} & \textbf{SC-IoU(\%)} & \textbf{SSC-mIoU(\%)} \\
\hline\hline
TSDF+RGB & \cmark &  &   & 83.1 & 52.5\\
TSDF+RGB &  &  \cmark &   & 82.6 & 53.2\\
TSDF+RGB &  &  &  \cmark  & \textbf{84.2} & \textbf{55.2}\\
\hline
\end{tabular}
}
\vspace{-0.35cm}
\end{center}
\caption{\textbf{Ablation studies on different representations of structure prior}. We perform this ablation study on NYUCAD dataset.}
\label{tab:ablation-prior}
\end{table}

\noindent \textbf{Different Representations of Structure Prior.}
We also perform ablation studies on different representations of structure prior. We list three different representations of the prior here: shape, semantic labels and sketch. Shape is the binary description of the scene and we generate the ground-truth shape by binarizing the semantic labels. Semantic labels and sketch have been introduced in the above sections. From Table \ref{tab:ablation-prior}, we observe that sketch is the best representation for modelling structure prior as it could infer the invisible regions with well structure-preserving details.

\begin{table}[t]
\begin{center}
\resizebox{\columnwidth}{!}{
\begin{tabular}{|c|c|c|c|} 
\hline
\textbf{Supervision} &  \textbf{Embedding} & \textbf{SC-IoU(\%)} & \textbf{SSC-mIoU(\%)} \\
\hline\hline
None & Implicit & 81.1 & 50.6\\
\hline
\multirow{2}{*}{Shape} & Implicit & 83.1 & 51.8\\
 & Explicit & \textbf{83.1} & \textbf{52.5} \\
 \hline
\multirow{2}{*}{Semantic} & Implicit & 82.3 & 52.1 \\
 & Explicit & \textbf{82.6} & \textbf{53.2} \\
 \hline
\multirow{2}{*}{Sketch} & Implicit & 83.5 & 54.4 \\
 & Explicit & \textbf{84.2} & \textbf{55.2} \\
\hline
\end{tabular}
}
\vspace{-0.35cm}
\end{center}
\caption{\textbf{Ablation studies on different types of embeddings}. We perform this ablation study on NYUCAD dataset.}
\label{tab:ablation-embedding}
\end{table}
\noindent \textbf{Different Types of Embeddings.} In this part, we conduct ablation studies on different types of embeddings. Results are shown in Table \ref{tab:ablation-embedding}. `Implicit' represents taking the output of the last deconvolution layer in the first stage as the geometric embedding and feed it to the second stage as prior information. `Explicit' represents we abstract a concrete structure based on the implicit embedding and use it a structure prior. We observe that even using implicit embedding, adding any reasonable supervision on it could boost the performance, such as semantics, shape and sketch. When we convert to explicit embedding, a better structure prior is obtained and the performance shows another boost. Note that the explicit embedding supervised by sketch outperforms its baseline using implicit embedding with no supervision by $3.1\%$ SC IoU and $4.6\%$ SSC mIoU, which demonstrates the effectiveness of the proposed sketch structure prior and the explicit embedding method.

\begin{table}[t]
\begin{center}
\resizebox{\columnwidth}{!}{
\begin{tabular}{|c|c|c|c|} 
\hline
\textbf{Input for Stage1} & \textbf{Input for Stage2} & \textbf{SC-IoU(\%)} & \textbf{SSC-mIoU(\%)} \\
\hline\hline
RGB & RGB  & 68.0 & 40.0\\
RGB & TSDF   & 71.2 & 40.2\\
TSDF & TSDF  & \textbf{71.5} & 37.2\\
TSDF & RGB  & 71.3 & \textbf{41.1}\\
\hline
\end{tabular}
}
\vspace{-0.35cm}
\end{center}
\caption{\textbf{Ablation studies on different modal input}. We perform this ablation study on NYU dataset.}
\label{tab:ablation-input}
\end{table}

\begin{figure*}[ht]
\centering
\includegraphics[width=\textwidth]{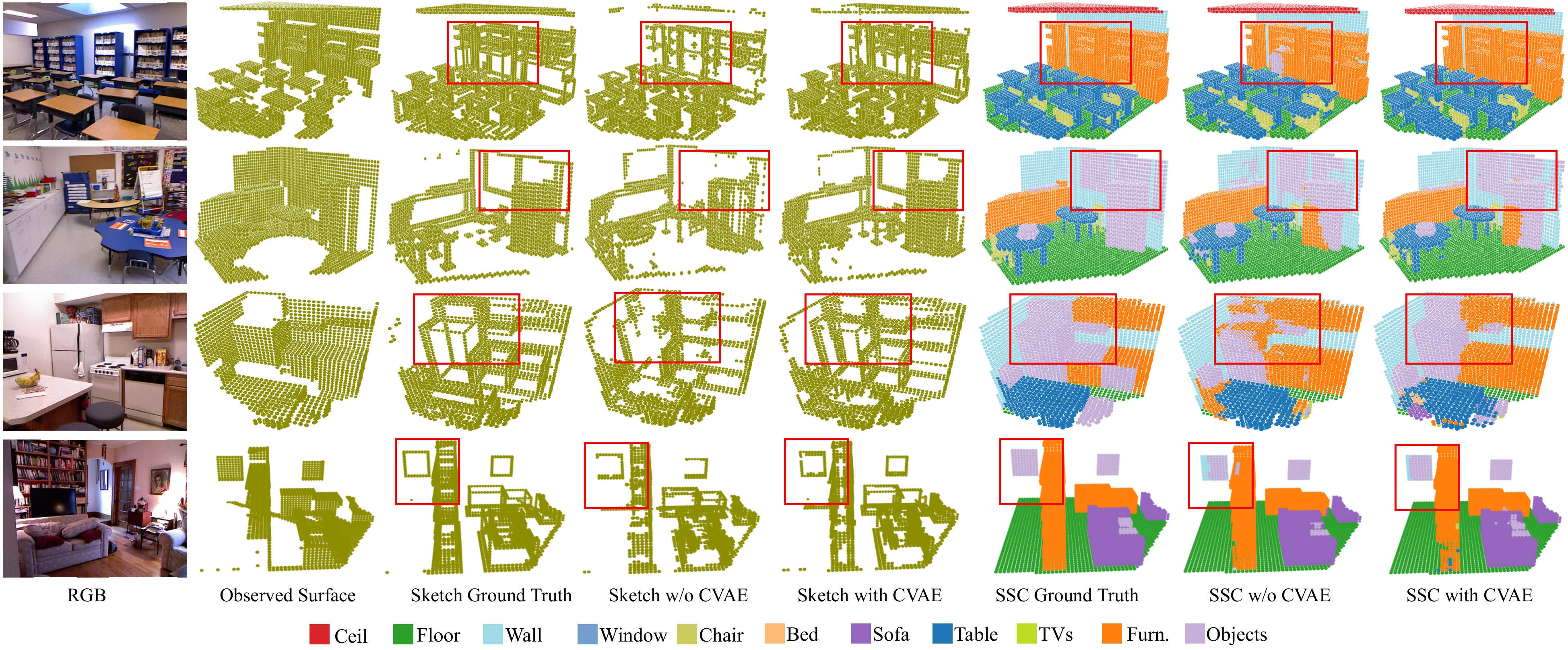}
\caption{\textbf{Visualization of the sketch on NYUCAD dataset.} With the proposed 3D Sketch Hallucination Module, which leverages CVAE to guide the inference of invisible areas, the sketch obtains a sharper boundary and is completer, resulting in better semantic predictions.}
\label{fig:nyucad-cvae}
\end{figure*}

\noindent \textbf{Different Modal Input.} We adopt data from different modalities as input, more specifically, TSDF for the first stage and RGB for the second stage. We claim that TSDF embeds rich geometric information and is suitable for the sketch prediction task, while RGB is rich in semantic information and is suitable for semantic label prediction task. Results are shown in Table \ref{tab:ablation-input}. From Row 1 and Row 4, we observe that TSDF generates better structure prior than RGB, resulting in a gain of $3.3\%$ SC IoU. From Row 3 and Row 4, we observe that RGB generates more precise semantic labels based on the same structure prior provided by TSDF, resulting in a gain of $3.9\%$ SSC mIoU. From Row 1, Row 2 and Row 3, we observe that the introduction of other modalities would result in corresponding gains on the basis of single-mode data.

\begin{table}[t]
\begin{center}
\resizebox{0.8\columnwidth}{!}{
\begin{tabular}{|c|c|c|c|} 
\hline
\textbf{Dateset} & \textbf{Resolution} & \textbf{SC-IoU(\%)} & \textbf{SSC-mIoU(\%)} \\
\hline\hline
NYU & (60, 60) & 71.3 & 41.1 \\
NYU & (80, 60) & 71.4 & \textbf{41.2} \\
NYU & (80, 80) & \textbf{76.5} & 40.0 \\
\hline
NYUCAD & (60, 60) & 84.2 & 55.2\\
NYUCAD & (80, 60) & 84.1 & \textbf{55.9}\\
NYUCAD & (80, 80) & \textbf{86.0} & 54.9 \\
\hline
\end{tabular}
}
\vspace{-0.2cm}
\end{center}
\caption{\textbf{Ablation studies on input/output resolutions}. We perform this ablation study on NYU and NYUCAD dataset both. \textit{Resolution(a, b)} means the input resolution is $(a \times 0.6a \times a)$ and the output resolution is $(b \times 0.6b \times b)$.}
\label{tab:ablation-resolution}
\end{table}

\noindent \textbf{Different Input/Output Resolutions.} In this part, we conduct ablation studies to verify the impacts of different input/output resolutions on the performance. Results are shown in Table \ref{tab:ablation-resolution}. We observe that increasing input size would not make the performance worse. If we increase both the input and output resolutions, SC IoU increases substantially, while SSC mIoU only declines slightly. Hence we conclude that increasing resolution of either input or output is beneficial to semantic scene completion task.

\subsection{Qualitative Results of 3D Sketch}
We visualize the predicted 3D sketch with/without CVAE in Figure \ref{fig:nyucad-cvae}. We can observe that the sketch is more complete and precise with the proposed 3D Sketch Hallucination Module. Under the constraints of a more complete sketch, the semantic result shows great consistency in regions with the same semantic labels and has a sharper boundary. For example, in the first row, some regions in the bookcase are mislabeled as $objects$ without CVAE, and those regions in the corresponding sketch are missing. In the second row, the sketch without CVAE fails to extract the outline of the object on the wall, leading to uncertainty of the semantic boundary. In the third row, the missing boundary in the sketch without CVAE brings confusing semantics. In the last row, the sketch of the photo frame is incomplete without CVAE, resulting in more areas to be mislabeled as $wall$.

\section{Conclusion}
In this paper, we propose a novel 3D sketch-aware feature embedding scheme which explicitly embeds geometric information with structure-preserving details. Based on this, we further propose a semantic scene completion framework that incorporates a novel 3D Sketch Hallucination Module to guide full 3D sketch inference from partial observation via structure prior. Experiments show the effectiveness and efficiency of the proposed method, and state-of-the-art performances on three public benchmarks are achieved.

\quad

\noindent \textbf{Acknowledgments:} This work is supported by the National Key Research and Development Program of China (2017YFB1002601, 2016QY02D0304), National Natural Science Foundation of China (61375022, 61403005, 61632003), Beijing Advanced Innovation Center for Intelligent Robots and Systems (2018IRS11), and PEK-SenseTime Joint Laboratory of Machine Vision.

{\small
\bibliographystyle{ieee_fullname}
\bibliography{egbib}
}

\end{document}